\documentclass[11pt]{article}

\usepackage{acl}

\usepackage{times}
\usepackage{latexsym}

\usepackage{tcolorbox}

\usepackage[T1]{fontenc}

\usepackage[utf8]{inputenc}

\usepackage{color}
\usepackage{tabularx}
\usepackage{tcolorbox}
\usepackage{xcolor}
\usepackage{colortbl}
\usepackage{tabu}
\usepackage{booktabs}
\usepackage{multirow}

\usepackage{graphicx} 
\usepackage{float}
\usepackage{subfigure} 
\usepackage{amssymb}
\usepackage{makecell}

\usepackage{amsmath}
\usepackage{hyperref}
\usepackage{tablefootnote}
\usepackage{xspace}
\usepackage{cleveref}
\crefrangelabelformat{section}{#3#1#4--#5\crefstripprefix{#1}{#2}#6}

\usepackage{float}
\usepackage{bm}
\usepackage{algorithmicx}
\usepackage{algorithm}
\usepackage{algpseudocode}
\usepackage{arydshln}

\usepackage{amssymb}
\usepackage{pifont}
\usepackage{enumitem}
\usepackage{makecell}
\usepackage{lipsum}

\newcommand{\eg}{\hbox{\emph{e.g.,}}\xspace}
\newcommand{\ie}{\hbox{\emph{i.e.}}\xspace}
\newcommand{\popqa}{\textsc{PopQA}\xspace}
\newcommand{\entityqa}{\textsc{EntityQuestions}\xspace}
\newcommand{\ours}{\textsc{RPDR}\xspace}

\title{\ours: A Round-trip Prediction-Based Data Augmentation Framework for Long-Tail Question Answering}

\author{
Yiming Zhang$^{1}$\thanks{Work done when Yiming Zhang was visiting NYU Shanghai.
Correspondence: Yiming Zhang (\texttt{yimingz@zju.edu.cn}), Chen Zhao (\texttt{cz1285@nyu.edu})} \quad
  Siyue Zhang$^{2}$ \quad
  Junbo Zhao$^{1}$ \quad
  Chen Zhao$^{3, 4}$ 
  \vspace{5pt}
  \\
  $^1$Zhejiang University \quad $^2$Nanyang Technological University \quad $^3$ NYU Shanghai \\ $^4$ Center for Data Science, New York University \vspace{3pt}\\
}

\begin{document} 
\maketitle
\begin{abstract}
Long-tail question answering presents significant challenges for large language models (LLMs) due to their limited ability to acquire and accurately recall less common knowledge. Retrieval-augmented generation (RAG) systems have shown great promise in mitigating this limitation by integrating external retrieval mechanisms. However, dense retrieval models often face the same difficulties when generalizing to rare or niche knowledge. In this study, we introduce \ours, a novel data augmentation framework that selects high-quality easy-to-learn training data, to enhance dense retrievers. Our approach is built around three core components: synthetic data generation, data selection with Round-Trip prediction to identify easy-to-learn instances, 
and retriever training with these instances. We evaluate \ours on two long-tail retrieval benchmarks, \popqa and \entityqa, demonstrating substantial improvements over existing retrievers like BM25 and Contriver, especially on extremely long-tail categories. We identify the strengths and limitations of \ours through detailed human analysis and propose a dynamic routing mechanism to dynamically route queries to specialized retrieval modules to further improve retrieval performance.\footnote{Our code will be made available at \url{https://github.com/yiming-zh/RPDR}.}

\end{abstract}

\section{Introduction}

\begin{figure}[t!]
    \centering
    \includegraphics[width=0.48\textwidth]{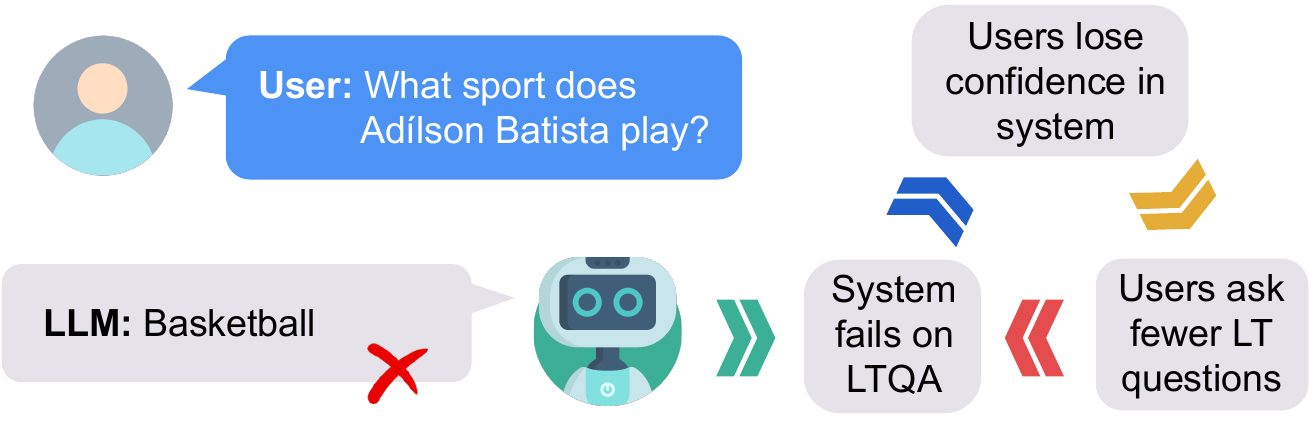}
    
    \vspace{0.05cm}
    
    \includegraphics[width=0.48\textwidth]{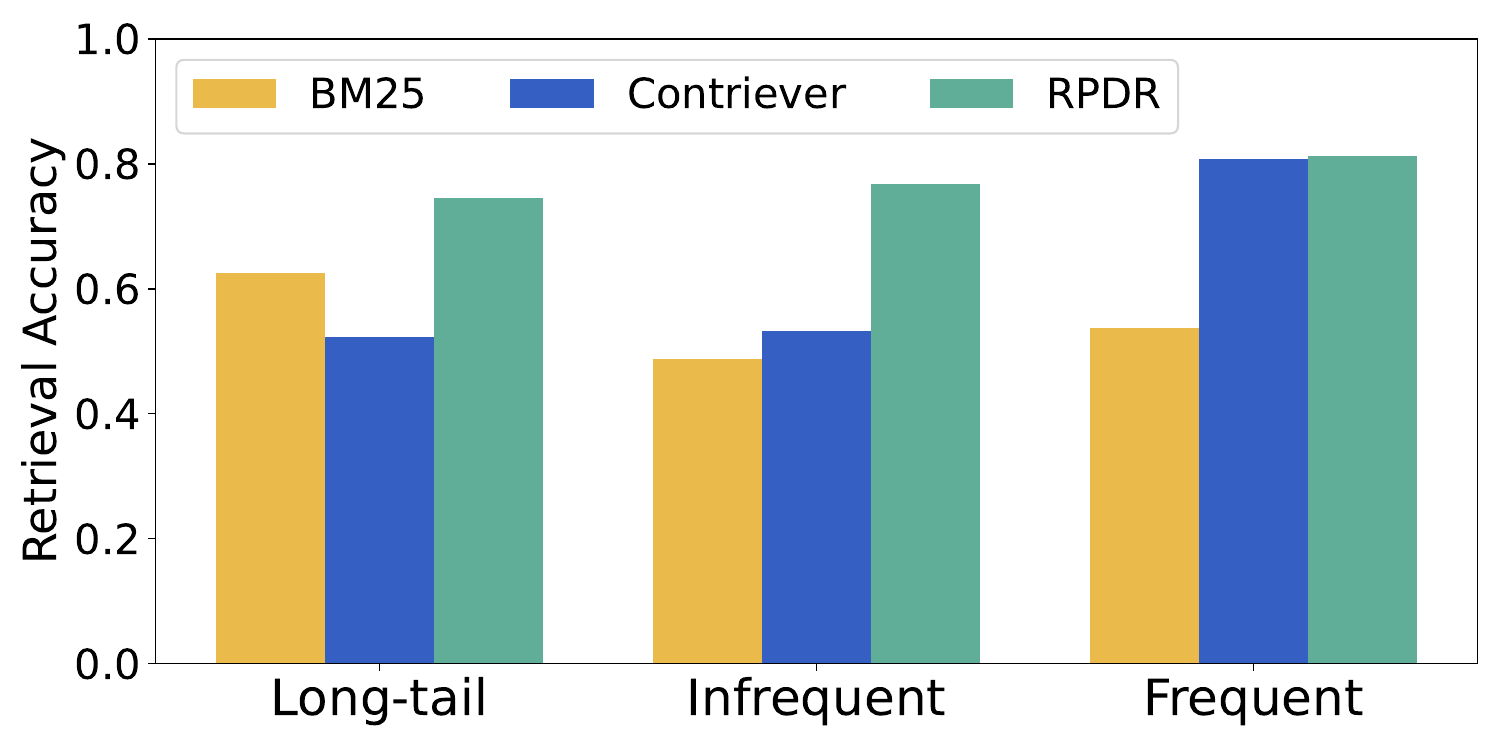}
    
    \captionsetup{justification=justified, singlelinecheck=false}  
    \caption{ (Top) Negative feedback loop in long-tail question answering in large language model-based systems. (Bottom) We challenge existing findings that dense retrievers struggle on long-tail questions, and argue that through appropriate training, dense retrieval based methods \ours can surpass BM25 on long-tail retrieval.}
    \label{fig: negative_loop}
\end{figure}

The development of large language models (LLMs) has transformed a wide range of applications, from answering questions to supporting complex decision-making \citep{rag_survey}. Despite their impressive capabilities, a key challenge arises when generalizing these models to real-world scenarios: handling long-tail user queries \citep{kandpal2023large, wild_lt}. For instance, a user might ask about a lesser-known football player as shown in \Cref{fig: negative_loop}. Addressing long-tail question answering (LTQA) is critical because it directly influences user trust and engagement. When systems fail to provide accurate answers to these uncommon queries, users may lose confidence in the system, reducing their likelihood of asking similar questions in the future. This interaction creates a negative feedback loop resulting in a reduction in long-tail training data that further degrades the model's performance.

Given the scale of today’s pre-training data and LLMs, it is expected that LLMs can acquire a huge amount of information and answer long-tail questions with their parametric knowledge. However, recent work ~\cite{kandpal2023large} has shown that LLMs struggle to learn long-tail knowledge,  and are prone to hallucinating in such cases. Retrieval-augmented generation (RAG) has emerged as a promising solution for LTQA \citep{rag_survey}. By integrating external knowledge through retrieval mechanisms, RAG systems aim to mitigate the limitations of relying solely on the parametric knowledge of LLMs. Despite their high potential, the technical challenge is how to retrieve relevant knowledge from millions of candidates for long-tail questions. Traditional term-matching-based information retrieval methods such as BM25 \citep{bm25}, fail to capture nuanced semantic similarities. Dense retrieval models ~\cite{karpukhin2020dense, contriever} address this by encoding both queries and knowledge pieces into a shared embedding space, allowing similarity to be computed through vector comparisons. However,  existing work shows that dense retrievers cannot generalize effectively to long-tail queries and perform even worse than traditional retrieval methods \citep{kandpal2023large, EntityQuestions}.

Our study challenges existing findings and claims that with appropriate training, dense retrieval methods can also excel on long-tail queries. Our intuition is to select easy-to-learn queries that are inherently learnable, in which the original text can be easily recovered from learned embeddings. To this end, we propose a novel data 
augmentation method \ours as follows: 
(1) \textbf{Synthetic Data Generation}, which generates new question-answer pairs, in which the questions include long-tail entities; (2) \textbf{Data Selection with Round-trip Prediction}, which filters and selects easy-to-learn question-answer pairs decided through a round-trip prediction: an off-the-shelf dense retriever \citep{contriever} produces an embedding, and the data is selected if a trained inverse model (a decoder) \citep{vec2text} can reconstruct the instance. (3) \textbf{Retriever Training with Augmented Data}, which trains a new dense retriever with selected easy-to-learn data.

We evaluate \ours on two long-tail question answering datasets: \popqa \citep{popqa} and \entityqa \citep{EntityQuestions}. \ours demonstrates significant improvements over BM25 and dense retrievers across long-tail and frequent query splits.  When combined with an LLM as a generator, the improved retrieval leads to higher answer accuracy, achieving 10.9\% improvement on long-tail queries.

Our analyses further indicate that using round-trip prediction to select augmented data substantially improves on a specific category that is common in long-tail queries: syntactically simple but semantically rare and nuanced entities (\eg \underline{John XIX} and \underline{John X}), while still struggling on syntactically complex entities (\eg \underline{Ernő Noskó}). Following these findings, we propose a routing mechanism to dynamically assign questions to different retrieval modules (\ie \ours or BM25) based on their characteristics. Compared with using only \ours, the whole framework achieves a 4.6\% improvement.

Our contributions are summarized as follows:
\begin{itemize}
    \item We challenge existing findings on retrieval-augmented generation approaches for long-tail question answering and claim that, when trained with augmented data, dense retrievers ultimately outperform traditional term-matching methods such as BM25. 
    \item We propose a new data augmentation framework \ours in the long-tail scenario, which selects easy-to-learn data through round-trip predictions.
    \item We provide a comprehensive study on the strengths and limitations of \ours and further propose a routing mechanism to mitigate these limitations.
\end{itemize}

\section{Background}
In this section, we first define the long-tail question answering task (\S\ref{sec:problem}), then introduce the baseline retrieval-augmented QA-based systems (\S\ref{sec:rag}), and finally, discuss the datasets used in our study (\S\ref{sec:dataset}).

\subsection{Long-Tail Question Answering}
\label{sec:problem}
We formulate our task as open-domain question answering \citep{karpukhin2020dense, chen2017reading}, where the goal is to take a question $q$ and predict an answer $y$. Our focus is on simple factoid questions, which rely on a single fact---represented as a triplet in the form of (subject, relation, object)---to derive the answer. Normally, the question contains the subject and relation (\eg ``what's the capital of the United States?''), and the answer is the corresponding object (\eg ``Washington D.C.''). We then define a question as long-tail if the frequency of the subject entity is smaller than a threshold, as follows:
\begin{equation}
f(e_q) \leq \tau,
\end{equation}
where \(\tau\) is the threshold distinguishing low-frequency entities from high-frequency ones.

With the development of LLMs, a straightforward approach is to directly ask these questions to an off-the-shelf LLM (\eg GPT-4o), with the hope that facts are captured in its parametric knowledge. However, answering long-tail questions poses a significant challenge in LLMs. As recent work~\cite{kandpal2023large} has shown even scaled-up LLMs still struggle to memorize long-tail knowledge that rarely appears in the pre-training data. To make matters worse, LLMs are prone to mixing up long-tail facts with more popular ones stored in their parametric knowledge, often resulting in hallucinations when answering these questions~\cite{kang2024unfamiliar}. Our study adopts an alternative approach that tackles limited memorization in parametric knowledge of LLMs, retrieval-augmented generation (RAG), which we detail next.

\subsection{Retrieval-Augmented Generation (RAG)}
\label{sec:rag}

The goal of RAG is to mitigate the limitations in the parametric knowledge of LLMs \citep{rag_survey}. RAG involves a separate retrieval module to find relevant passages from a large corpus (\eg Wikipedia) and an LLM to generate an answer based on the question and retrieved passages.

\paragraph{Passage retrieval.} Retrieval methods are typically categorized into sparse retrieval and dense retrieval. Sparse retrieval methods, like TF-IDF \citep{tf_idf} and BM25 \citep{bm25}, rely on surface form matching. In contrast, dense retrieval methods encode questions and passages into embeddings, allowing semantic matching.

Specifically, dense retrieval models use PLMs (\eg BERT \citep{sentencebert}) to encode the question $q$ and the passage $p$ separately using two independent encoders (\ie bi-encoders \citep{karpukhin2020dense}). These models learn a scoring function (\eg dot product) between question and passage vectors:
\begin{equation}
    f(q,p) = \text{sim}(\text{Enc}_{Q}(q), \text{Enc}_{P}(p)).
\end{equation}
Dense retrievers are highly scalable, since passages
can be encoded offline, and are efficiently retrieved
over maximum inner product search (MIPS) with the question.

For training,  dense retriever models are primarily based on the contrastive learning paradigm. Specifically, given a positive passage $p^{+}$ and a set of negative passages $p_{1}^{-}, ..., p_{m}^{-}$, we use negative log-likelihood (NLL) loss:
\begin{equation}
  L(q, p^{+}, p_{1}^{-}, ..., p_{m}^{-})={\frac{e^{f(q, p^{+})}}{e^{f(q, p^{+})} + \sum_{j=1}^{m} {e^{f(q, p^{-}_{j})}}}}.
  \label{eqn:loss}
\end{equation}

\paragraph{Answer generation.} The answer generator takes as input the question, as well as the top passages from the retrieval component, and generates the answer. A widely used approach is Fusion-in-Decoder, which fine-tunes an encoder-decoder LLM (e.g., T5 ). FiD independently encodes each passage and question, then concatenates their representations before passing them to the decoder, as formulated below:
\begin{equation}
    y = \text{Dec}([\text{Enc}([q;p_1]);\dots;\text{Enc}([q;p_k])]), p_k \in \mathcal{D}.
\end{equation}
With decoder-only LLMs, the support passages and the question are concatenated into a single sequence as follows:
\begin{equation}
    y = \text{Dec}([p_1, \dots, p_k;q])), p_k \in \mathcal{D}.
\end{equation}

\begin{table}[t!]
\centering 
\small
\resizebox{\columnwidth}{!}{
\renewcommand{\arraystretch}{1.2}
\begin{tabular}{lrr}
\toprule
\textbf{Property}                            & \textbf{\popqa} & \textbf{\entityqa}  \\
\midrule
\multicolumn{3}{c}{\textbf{Knowledge Corpus}} \\
Source & Wikipedia & Wikipedia \\
\noalign{\vskip 0.5ex}\hdashline\noalign{\vskip 0.5ex}
\multicolumn{3}{c}{\textbf{Dataset Statistics}} \\
\# Relations    & 16 & 24 \\

Dataset Size (\# Questions)   &   &     \\
\quad \# Training Set & NA & 176.6k\\
\quad \# Development Set & NA & 22.1k\\
\quad \# Test Set & 14.3k & 22.1k\\
\toprule
\end{tabular}
}
\caption{Basic statistics of the \popqa and \entityqa used in our experiments. ``NA" means no such set.}
\label{tab:basic-stats}
\end{table}

\subsection{Dataset}
\label{sec:dataset}

Considering that the long-tail RAG scenario requires two conditions, long-tail knowledge and a supporting corpus, we conducted experiments and analyses on two benchmark datasets, \popqa \cite{popqa} and \entityqa \cite{EntityQuestions}, and their statistics are summarized in \Cref{tab:basic-stats}.

\begin{itemize}[leftmargin=*]
    \itemsep0em 

\item \textbf{\popqa} is a large-scale long-tail question answering dataset about entities with a wide variety of popularity. It is grounded in Wikidata. 
Beyond entities, \popqa features a diverse array of question types, covering topics such as people, places, organizations, and events.

\item \textbf{\entityqa} is another long-tail question-answering dataset. Similar to \popqa, it uses Wikipedia hyperlink counts as a proxy for the frequency of entities and samples knowledge triples from Wikidata. \entityqa includes seventeen different relation types. 
\end{itemize}

\section{Method}

\begin{figure*}[t!]
    \centering
    \includegraphics[width=0.95\textwidth]{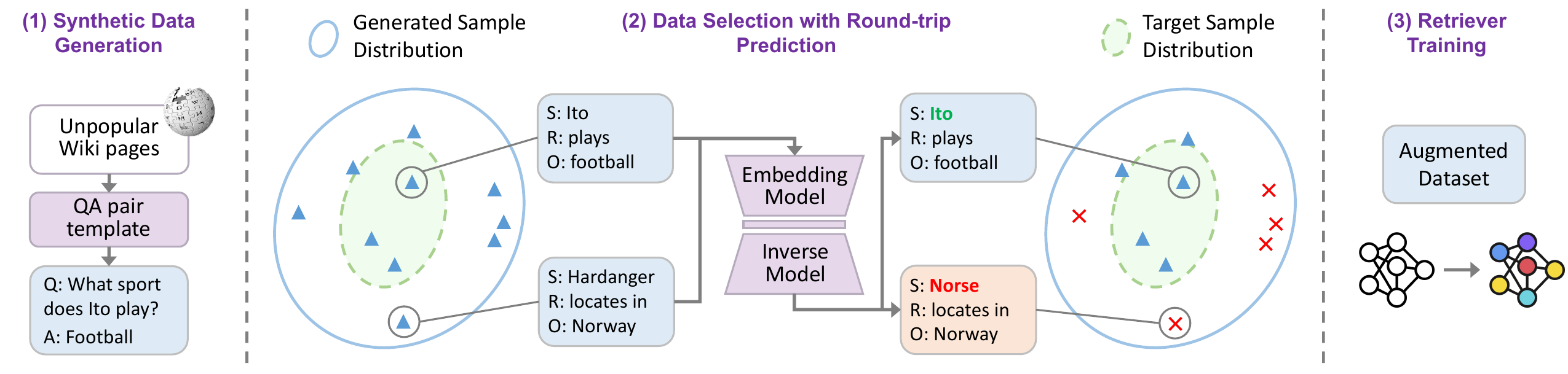}
    \captionsetup{justification=justified, singlelinecheck=false}  
    \caption{The RPDR framework consists of three main stages: (1) Synthetic Data Generation that generates a pool of long-tail QA pairs. (2) Data Selection with Round-trip Prediction that trains an inverse model to select easy-to-learn samples with reversibility. (3) Retriever Training that trains a dense retriever using the augmented samples. 
    }
    \label{fig:system}
\end{figure*}
This section introduces our proposed framework, \ours.
As mentioned earlier, we focus on enhancing the capability of dense retrievers for long-tail entities---a challenge that significantly impacts the performance of Retrieval-Augmented Generation (RAG) systems. At a high level, as illustrated in \Cref{fig:system}, \ours employs a data augmentation approach, generating new synthetic samples containing long-tail entities to train dense retrievers (\S\ref{sec:da}). We emphasize that the quality of these augmented samples is crucial, ensuring that their patterns are easier for dense retrievers to learn.
We further hypothesize that easy-to-learn data yield embeddings that can be reliably recovered, as the mapping between text and embeddings is inherently learnable. To this end, we introduce a novel round-trip prediction mechanism to identify high-quality, easy-to-learn data. Specifically, we first train a separate inverse embedding model (\S\ref{sec:inv_emb}). Given a sample, we use an off-the-shelf dense retriever to generate embeddings from the question and select the sample for training a new dense retriever only if its embedding can be accurately reconstructed by the inverse embedding model (\S\ref{sec:data_select}).

\subsection{Synthetic Data Generation}
\label{sec:da}

We adopt similar mechanisms to \popqa \citep{popqa} and \entityqa \citep{EntityQuestions} to create the pool from Wikipedia for data augmentation. Specifically, we synthesize samples using the knowledge triples from Wikidata. To ensure that the selected samples are in the long-tail distribution, we use the monthly page view count of the Wikipedia page corresponding to the subject of the knowledge triple answer as a measure of each sample's popularity. We only keep triples with popularity below a predefined threshold.\footnote{In this paper, we set the threshold of popularity at 1e6. It is also worth noting that during data augmentation, we remove triples already used in \popqa while ensuring that the entire process remains invisible to the \entityqa test set.}
Then, for each triple, we apply templates to generate query-answer pairs, where each query is about the subject and relation, and the answer is the corresponding object. Next, we adopt BM25 to retrieve the top 1,000 passages. We only keep the sample if these passages contain the correct answer string. In total, we generated approximately 86k samples for \popqa and approximately 126k samples for \entityqa.

\subsection{Training an Inverse Model}
\label{sec:inv_emb}
The inverse model \cite{vec2text} is designed to reconstruct original text from an embedding. Specifically, given an embedding $e$ generated by a specified embedding model, our goal is to recover the corresponding text $x$. This can be formulated as generating text whose embedding is maximally similar to the ground-truth embedding as:
\begin{equation}
\hat{x} = \arg\max_{x} \cos(\phi(x), e).
\end{equation}

Given a dataset $\mathcal{D} = \{x_1, \dots\, x_n\}$, we train an 
inverse mapping of the encoder  $\phi$ by modeling the conditional distribution of text given their embeddings $p(x \mid e; \theta)$. We achieve this by optimizing $\theta$ via maximum likelihood estimation:
\begin{equation}
\theta = \arg \max_{\hat{\theta}} \mathbb{E}_{x \sim \mathcal{D}} \big[ p(x \mid \phi(x); \hat{\theta}) \big].
\end{equation}
For technical details, readers can refer to \citet{vec2text} for more information.

\subsection{Data Selection with Round-trip Prediction}
\label{sec:data_select}

After training the inverse model, we select easy-to-learn augmented data for training dense retriever models. For each question-answer sample, we first apply an off-the-shelf dense retriever \citep{contriever} on the question $x$ to get an embedding $e(x)$, then we adopt the trained inverse embedding model for the reconstructed text $\hat{x} $. Specifically, we define a reconstruction quality score $ S(x) $ for $ x $ as follows:
\begin{equation}
S(x) = 1 - \frac{\| \phi(\hat{x}) - \phi(x) \|^2}{\| \phi(x) \|^2}.
\end{equation}
A higher value of $ S(x) $ indicates better reconstruction quality. We then select the top $k$ samples in an augmented set $D_{\text{selected}}$, where the reconstruction score $ S(x) $ is above a threshold $ \theta $:
\begin{equation}
D_{\text{selected}} = \{ x \mid S(x) \geq \theta, \, x \in D_{\text{aug}} \}.
\end{equation}

\label{sec:train_retriever}
With the filtered dataset \(D_{\text{selected}}\), we use the contrastive learning objective mentioned in \Cref{eqn:loss} to fine-tune the off-the-shelf dense retriever, Contriver \cite{contriever}.

\section{Experiment}
In this section, we evaluate \ours and baseline systems on \popqa and \entityqa.

\begin{table*}[htbp]
\centering
\renewcommand{\arraystretch}{1.3} 
\setlength{\tabcolsep}{10pt}      
\resizebox{\textwidth}{!}{
\begin{tabular}{lcccccccccccc}
\toprule
\multicolumn{1}{c}{\multirow{3}{*}{\textbf{Method}}}  & \multicolumn{6}{c}{\textbf{\popqa}}         & \multicolumn{6}{c}{\textbf{\entityqa}}         \\
\cmidrule(lr){2-7} \cmidrule(lr){8-13} 
\multicolumn{1}{c}{}  & \multicolumn{2}{c}{\textbf{Long-tail}} & \multicolumn{2}{c}{\textbf{Infrequent}} & \multicolumn{2}{c}{\textbf{Frequent}} & \multicolumn{2}{c}{\textbf{Long-tail}} & \multicolumn{2}{c}{\textbf{Infrequent}} & \multicolumn{2}{c}{\textbf{Frequent}} \\
\cmidrule(lr){2-3} \cmidrule(lr){4-5} \cmidrule(lr){6-7} \cmidrule(lr){8-9} \cmidrule(lr){10-11} \cmidrule(lr){12-13}
\multicolumn{1}{c}{}  & R@10 & R@20 & R@10 & R@20 & R@10 & R@20 & R@10 & R@20 & R@10 & R@20 & R@10 & R@20 \\
\hline
BM 25     & 62.6 & 66.8 & 46.2 & 52.8  & 53.8 & 61.1  & 48.3 & 63.5 & 41.9 & 66.7  & 55.0 & 69.2  \\
Contriever       & 52.3 & 59.3 & 53.3 & 61.1  & 80.0 & 85.8  & 42.9 & 53.3 & 53.4 & 61.0  & 57.3 & 67.6  \\
BGE             & 53.9 & 61.2  & 57.4 & 65.3 & 82.4 & 87.9 & 44.7 & 60.3 & 55.1 & 69.8 & 60.9  & 75.2 \\
NV-Embed              & 55.0 & 62.3  & 63.5 & 70.6 & \textbf{89.9} & \textbf{90.6} & 48.4 & 61.6 & 57.1 & 70.7 & \textbf{71.5} & \textbf{84.3} \\
Gemma             & 53.8 & 61.7  & 66.9 & 69.3 & 88.7 & 90.2 & 49.3 & 61.1 & \textbf{58.2} & \textbf{71.5} & 70.5 & 82.7 \\

\hline
\ours               & \textbf{74.5} & \textbf{75.7} & \textbf{76.8} & \textbf{78.2}  & 81.3 & 87.1  & \textbf{54.6} & \textbf{68.1} & 57.4 & 71.2  & 62.7 & 70.0  \\
\ours-Random                 & 66.8 & 70.0 & 70.0 & 74.5  & 82.7 & 87.3  & 51.9 & 67.7 & 52.4 & 69.5  & 55.9 & 70.8  \\
\bottomrule
\end{tabular}
}

\captionsetup{justification=justified,singlelinecheck=false}

\caption{Comparison of retrieval performance between RPDR and baseline retrievers for questions with long-tail, infrequent, and frequent entities in the \popqa and \entityqa datasets. ``R@k'' means Retrieval Accuracy in the top-k retrieved passages. ``RPDR-random'' is a model trained with randomly sampled data from the augmented long-tail distribution. More details refer to \Cref{appendix:random}.
}
\label{tab:performance_comparison}
\end{table*}

\subsection{Experimental Setup}
\paragraph{Baselines.} 
We selected two basic retrieval methods as baselines for experimental comparison: the text-based approach BM25 \citep{bm25} and the widely used embedding-based approach Contriever \citep{contriever}. Besides, we selected three state-of-the-art embedding-based retriever models, BGE \citep{bge_m3}, gemma \citep{gemma}, and NV-Embed \citep{lee2024nv}. We include details in  \Cref{appendix: details of experiments}. 

\paragraph{Metrics.}
Following \citep{popqa,wang2025proprag,yu2024rankrag}, for retrieval evaluation, we mainly use Retrieval Accuracy, denoted as \textbf{R@k}, which measures whether at least one answer string appears in the top-\(k\) retrieved passages. For answer evaluation, we use \textbf{Exact Match}, which measures the exact string match between the predicted and the gold answer.

To provide a systematic study, we partition the \popqa and \entityqa test set with different entity frequencies. Specifically, we split dataset into three categories: \textbf{Long-Tail}  that represents queries with corresponding entity frequencies between 10 and 100, capturing extreme rare cases; \textbf{Infrequent} that corresponds to entity frequencies between 100 and 10,000, offering a moderately rare queries; \textbf{Frequent} that with frequencies exceeding 10,000, correspond to high-frequency queries involving well-known entities.

 \paragraph{Training Details.} 
We follow \citet{vec2text}'s setting to train inverse models. We use an off-the-shelf Contriever as an encoder and train the T5-base model on MS MARCO dataset \citep{marco}, with a batch size of 128, a maximum of 100 epochs with early stopping, and the Adam optimizer~\cite{adam} with a learning rate of 1e-3. 
For \ours, we fine-tune the Contriver model using the selected samples, with a batch size of 32 for 15 epochs, and the AdamW optimizer with an initial learning rate of 5e-6. 
For \popqa, we include 22k augmented samples as training data. For \entityqa, we combine 41K augmented samples with its original training set (176.6K) as our training data. 
We prompt LLMs with 15 examples for answer generation, following \citet{popqa}.

\subsection{Main Results}

\paragraph{BM25 excels on long-tail queries while dense retrievers outperform other types.}
Consistent with previous work ~\cite{popqa, EntityQuestions}, as shown in Table~\ref{tab:performance_comparison}, we find that BM25 outperforms all dense retrievers on long-tail queries.  This is because dense retrieval models struggle to generalize to represent long-tail entities. As expected, dense retrievers outperform BM25 in frequent and infrequent queries.

\paragraph{\ours significantly enhances long-tail retrieval.}
According to Table~\ref{tab:performance_comparison}, \ours significantly outperforms all other methods on long-tail queries, highlighting the crucial role of data augmentation. For example, in the long-tail category of the PopQA dataset, \ours achieves a 19.5\% improvement in R@10 over NV-Embed, the best-performing embedding model trained from larger decoder-only LLMs. \textbf{Notably, with data augmentation, \ours also surpasses BM25 by 11.9\% in R@10, further supporting our claim that, with appropriate training, dense retrieval methods can excel in long-tail query scenarios}. On frequent categories, as expected, larger embedding models achieve better performance, while \ours also slightly outperforms Contriever, which is also initialized from a BERT model. \footnote{Due to limited computational resources, we initialized \ours from a Contriever model. We anticipate that starting from decoder-only LLMs (e.g., LLaMA) would yield additional improvements on both long-tail and other categories, which we leave for future work.}

\paragraph{Data augmentation with round-trip prediction is a key factor.}
The comparison between \ours and \ours-Random in \Cref{tab:performance_comparison} indicates the benefits of round-trip prediction-based data augmentation. For instance, in the long-tail category of PopQA, \ours outperforms \ours-Random by 7.7\%. These improvements validate our intuition that the proposed data selection scheme effectively identifies easy-to-learn data, resulting in more efficient learning.

\paragraph{The improvements in retrieval propagate to QA accuracy.}
According to Table \ref{tab:model_comparison}, relying solely on the parametric knowledge of LLMs proves ineffective for long-tail queries. Retrieval augmentation mitigates this issue, and with the superior retrieval capabilities of \ours, answer accuracy improves significantly over baseline methods. For instance, when using GPT-3.5 as a generator, \ours surpasses Contriever as a retriever by 11.7\%.

\begin{table*}[htbp]
\centering
\renewcommand{\arraystretch}{1.1} 
\setlength{\tabcolsep}{10pt}      
\resizebox{\textwidth}{!}{
\begin{tabular}{llcccccc}
\toprule
\multicolumn{1}{c}{\multirow{2}{*}{\textbf{Generator}}} & \multicolumn{1}{c}{\multirow{2}{*}{\textbf{Retriever}}} & \multicolumn{3}{c}{\textbf{\popqa}}   & \multicolumn{3}{c}{\textbf{\entityqa}}                 \\ 
\cmidrule(lr){3-5} \cmidrule(lr){6-8}
\multicolumn{2}{c}{}      & \textbf{Long-tail} & \textbf{Infrequent} & \textbf{Frequent} & \textbf{Long-tail} & \textbf{Infrequent} & \textbf{Frequent} \\ \hline
\multirow{4}{*}{GPT-j-6B}           &  Parametric         & 11.3                &   14.4             &      18.2         &    9.7               &       12.1         &        12.9           \\
                                    & BM25               & 24.9                &   19.4             &   24.5           &     23.2              &     27.1           &      29.8          \\
                                    & Contriever         & 19.2                &   24.7               &     39.1         &   16.5                 &     26.6           &     30.2        \\
                                    & \ours               & 30.2                &   29.7             &      40.0         &    28.9             &           32.8     &        29.2       \\ \hline
\multirow{4}{*}{GPT-neox 20B}       &  Parametric        & 13.5                &   21.2              &     28.9         &    10.1               &       12.8          &      14.8           \\
                                    & BM25               & 29.1                &   27.4               &      29.2        &    27.8            &            29.9     &       35.6      \\
                                    & Contriever         & 25.6                &   33.2             &      40.7         &     23.7          &      24.9           &        34.4         \\
                                    & \ours               & 34.9                &   35.5             &      40.0          &    33.5                &      33.7           &     35.9          \\ \hline
\multirow{4}{*}{LLaMA-3 8B}       &  Parametric        & 11.9                &   21.6              &     29.9         &    10.3               &       14.5          &      21.7           \\
                                    & BM25               & 24.7                &   29.2               &      30.1        &    27.8            &            29.6     &       32.8      \\
                                    & Contriever         & 23.7                &   34.9             &      39.1         &     21.8          &      25.4           &        30.2         \\
                                    & \ours               & 29.8                &   34.7             &      41.8          &    30.1                &      31.6           &     33.9          \\ \hline
\multirow{4}{*}{GPT-3.5}            &  Parametric         & 22.7                &   36.4             &     50.3           &   20.6                &     29.9            &     36.8 \\
                                    & BM25               & 30.5                &   31.8             &     42.6           &   35.9               &      35.4             &    35.6               \\
                                    & Contriever         & 29.7                &   41.5             &      52.7           &    32.5             &        37.8           &        38.4          \\
                                    & \ours               & 41.4                &   42.9              &  52.1             &    36.4             &       38.6             &    38.7              \\ 
\bottomrule
\end{tabular}}
\captionsetup{justification=justified,singlelinecheck=false}

\caption{End-to-end QA exact match accuracy comparison for RAG systems with different retriever and reader models on the \popqa and the \entityqa datasets. ``Parametric'' means the generator models answer the query by their parametric knowledge.}
\label{tab:model_comparison}
\end{table*}

\subsection{Ablation Study}

\ours augments training data only when both the question and the answer are reconstructed successfully. To analyze the impact of different data selection criteria, we conduct an ablation study. Specifically, based on the reconstruction accuracy of the question entity (Q) and answer entity (A), we categorize the data into three groups and train retrievers accordingly: (1) Q correct \& A correct, (2) Q correct \& A wrong, and (4) Q wrong \& A wrong.\footnote{During data generation, we ensure that the query entity has low frequency while the answer does not. As a result, there are very few samples for the ``Q correct \& A wrong'' category, so we exclude it from our study.}

\paragraph{Training with incorrectly recovered samples hurts performance.} According to \Cref{tab:performance_ablation}, training on samples that cannot be recovered (including those with incorrect Q or A) negatively impacts retrieval performance. We hypothesize that training on long-tail queries inherently disrupts the patterns previously learned by the model. In contrast, samples that can be effectively recovered are easier to learn and help mitigate catastrophic forgetting ~\cite{khalifa2020distributional, korbak2022reinforcement}.

\begin{table}[htbp]
\centering
\renewcommand{\arraystretch}{1.1} 
\resizebox{1.0\linewidth}{!}{
\begin{tabular}{lcccccc}
\toprule
\multicolumn{1}{c}{\multirow{2}{*}{\textbf{Category}}}    & \multicolumn{2}{c}{\textbf{Long-tail}} & \multicolumn{2}{c}{\textbf{Infrequent}} & \multicolumn{2}{c}{\textbf{Frequent}} \\
\cmidrule(lr){2-3} \cmidrule(lr){4-5} \cmidrule(lr){6-7}
\multicolumn{1}{c}{}  & R@10 & R@20 & R@10 & R@20 & R@10 & R@20  \\
\hline
Q correct  \& A correct        &74.5 & 75.7 & 76.8 & 78.2  & 81.3 & 87.1    \\
Q wrong  \& A correct     & 66.8 & 74.9 &  68.6 & 76.6  & 77.6  & 85.4    \\
Q wrong \& A wrong      & 62.9 & 69.8 &  53.9 & 62.0  & 65.8  & 73.8    \\
\ours-Random                 & 66.8 & 70.0 & 70.0 & 74.5  & 82.7 & 87.3   \\
\bottomrule
\end{tabular}
}
\captionsetup{justification=justified,singlelinecheck=false}
\caption{Retrieval performance on the \popqa dataset for retrievers trained with different categories of augmented data. ``Q'' means the question entity, while ``A'' means the answer entity. ``correct''/``wrong'' means the entity can be recovered correctly/wrongly.}
\label{tab:performance_ablation}
\end{table}

\section{Qualitative Analysis of \ours and A Routing Mechanism}
This section first conducts qualitative analysis on the strengths and limitations of \ours (\S\ref{sec:qualitative}), motivated by these findings, we propose a routing mechanism that aggregates the strengths of multiple retrievers (\S\ref{sec: adaptive}). 

\subsection{Qualitative Analysis}
\label{sec:qualitative}

We qualitatively assess the strengths and weaknesses of \ours by manually reviewing 50 examples where \ours retrieves the correct passage while Contriever does not, as well as 50 cases where \ours fails to locate the correct passage. Additional examples are provided in \Cref{appendix: Case Study}.

\begin{figure}[t!]
    \centering
    \includegraphics[width=0.48\textwidth]{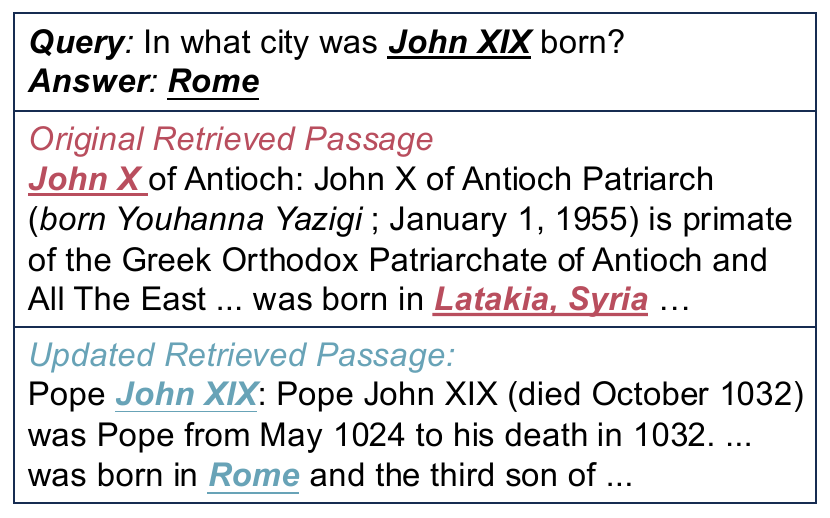}
    \captionsetup{justification=justified, singlelinecheck=false}  
    \caption{Case study of ``John XIX''. ``Original'' means the top-1 retrieved passage from the off-the-shelf Contriever, while ``Updated'' means the one from \ours.}
    \label{fig:case_study}
\end{figure}

\paragraph{\ours represents nuanced subword differences in long-tail entities.} 
As shown in \Cref{fig:case_study}, many long-tail entities differ only by small subword variations, such as \underline{John XIX} and \underline{John X}. Dense retrievers struggle with these cases because such entities are rare in the training data, causing them to primarily focus on matching the common word ``John''. By leveraging data augmentation with round-trip prediction, \ours learns these nuanced patterns and encodes the distinctions.

\paragraph{\ours (still) struggles with long-tail entities that have complex syntactic structures.} Through error analysis, we find that \ours cannot embed syntactically complex entities, in 72\% of error cases. These entities (\eg \underline{Ernő Noskó}) often contain complex morphological structures or rare characters, making them harder to learn, and are not included as augmented data (as our inverse embeddings also fail on them). We argue that in such cases, term-matching approaches like BM25 are more effective.

\paragraph{Questions in these datasets are ambiguous.}  We find that  24\% of errors stem from question ambiguity. For example, entities like \underline{Finale} can refer to either an album or a song, leading to multiple correct answers to the question, ``Who is the creator of Finale?''---yet the labels only include \underline{Madeon}. Additionally, 4\% of errors involve equivalent answers \citep{si2021s}. For instance, the retrieved content may contain the word ``novel'' while the expected answer is ``fiction''.

\subsection{Routing Mechanism}
\label{sec: adaptive}
Through qualitative analysis, we identify clear patterns in \ours's strengths and limitations. Based on these insights, we propose a routing mechanism that dynamically switches between two retrieval strategies---BM25 and \ours---depending on the characteristics of the input query ~\cite{popqa}. Specifically, we formulate this as a binary classification problem, where the input is the query, and the output is a binary label that determines which retriever to route to.

\paragraph{Setup.}  We fine-tune a Sentence-BERT \citep{sentencebert} model for classification. We randomly select 10,000 augmented samples as training data, with half correctly predicted by Contriever and the other half by BM25. We use a batch size of 32 for 5 epochs with the AdamW optimizer and an initial learning rate of 2e-5.

\paragraph{Results.} As shown in \Cref{tab:Routing Comparision}, 
our routing mechanism further improves retrieval performance, particularly for long-tail and infrequent categories. For instance, on \popqa, RPDR + RM achieves a 4.6\% improvement in R@10 over \ours. These results validate our hypothesis that combining the strengths of different retrievers is beneficial.

\begin{table}[htbp]
\centering
\renewcommand{\arraystretch}{1.3} 
\setlength{\tabcolsep}{10pt}      
\resizebox{0.5\textwidth}{!}{
\begin{tabular}{lcccccc}
\toprule
\multicolumn{7}{c}{\textbf{PopQA}} \\
\midrule
\multirow{2}{*}{\textbf{Method}}
& \multicolumn{2}{c}{\textbf{Long-tail}}
& \multicolumn{2}{c}{\textbf{Infrequency}}
& \multicolumn{2}{c}{\textbf{Frequency}} \\
\cmidrule(lr){2-3} \cmidrule(lr){4-5} \cmidrule(lr){6-7}
& R@10 & R@20 & R@10 & R@20 & R@10 & R@20 \\
\midrule
\ours        & 74.5 & 75.7 & 76.8 & 78.2 & 81.3 & 87.1 \\
\ours + RM   & 79.1 & 80.3 & 78.7 & 80.0 & 81.3 & 87.9 \\
\midrule
\multicolumn{7}{c}{\textbf{\entityqa}} \\
\midrule
\multirow{2}{*}{\textbf{Method}}
& \multicolumn{2}{c}{\textbf{Long-tail}}
& \multicolumn{2}{c}{\textbf{Infrequency}}
& \multicolumn{2}{c}{\textbf{Frequency}} \\
\cmidrule(lr){2-3} \cmidrule(lr){4-5} \cmidrule(lr){6-7}
& R@10 & R@20 & R@10 & R@20 & R@10 & R@20 \\
\midrule
\ours       & 54.6 & 68.1 & 57.4 & 71.2 & 62.7 & 70.0 \\
\ours + RM  & 58.4 & 72.2 & 59.5 & 71.3 & 61.8 & 71.1 \\
\bottomrule
\end{tabular}
}
\captionsetup{justification=justified,singlelinecheck=false}
\caption{Retrieval performance comparison of RPDR without and with the routing mechanism on the \popqa and the \entityqa datasets.}
\label{tab:Routing Comparision}
\end{table}

\section{Related Work}
\paragraph{LLM hallucinations.}
Large Language Models are susceptible to hallucinations, generating plausible yet incorrect content \citep{RAGTruth}. To address this, research has focused on retrieval-augmented methods that ground outputs in external knowledge sources \citep{asai2023self, siriwardhana2023improving, reichman2024dense}.  Our work builds on retrieval-augmented methods, seeking to enhance the generalization capabilities of LLMs in long-tail scenarios where existing methods still falter.

\paragraph{Dense retrieval models.} The retrieval model plays an important role in the retrieval-augmented generation \citep{fan2024survey}.  
Dense retrieval models encode both queries and documents into dense vectors using transformer-based models~\cite{karpukhin2020dense, contriever, xiong2021dense,chen2017hybrid}. These models outperform sparse retrieval methods like BM25 \citep{robertson2009probabilistic}, by capturing semantic similarity more effectively.
 
Recent studies have focused on distilling or fine-tuning LLM embeddings and demonstrated outstanding performance~\cite{bge_m3,gemma,hu2024minicpm,lee2024nv}. We focus on improving the dense retriever with data augmentation for long-tail question answering, where their performance used to lag behind.

\paragraph{Data augmentation.} Data augmentation~\cite{chen2023empirical} has emerged as a critical technique for addressing data scarcity in the NLP community.  There are various data augmentation approaches, from token level such as synonym replacement~\cite{kolomiyets2011model} to sentence level like back-translation~\cite{edunov2018understanding}. Recent approaches mainly sample LLMs to get augmented data~\cite{ding2024data}. We propose a round-trip prediction to select high-quality and easy-to-learn augmented data for long-tail question answering.

\section{Conclusion}
In this work, we challenge existing findings on long-tail QA by arguing that dense retrieval methods, when trained with high-quality and easy-to-learn augmented data, can outperform traditional term-matching approaches such as BM25. Our proposed \ours framework enhances retrieval quality by adopting synthetic data generation and a novel round-trip prediction mechanism, leading to significant improvements on benchmark datasets and boosting overall end-to-end system performance. Furthermore, our analysis sheds light on the strengths and limitations of \ours, and we introduce a routing mechanism that combines the advantages of multiple retrievers, offering additional performance enhancements.

\section{Limitations}
First, our round-trip prediction framework for selecting easy-to-learn augmented data incurs additional computational costs during data construction and model fine-tuning. 
Second, existing long-tail QA datasets primarily focus on single-fact questions, leaving the extension of our findings to long-tail questions requiring complex reasoning (e.g., multi-hop questions) largely unexplored. 
Third, while our work mainly focuses on short-form question answering, future research could extend this approach to long-form generation tasks that necessitate long-tail knowledge as part of the process.
At last, some valuable datasets, such as Head-to-Tail~\cite{sun2023head}, cannot serve as our experimental settings either due to the lack of an associated corpus or because they do not reflect the long-tail scenarios that our work targets.


\section{Acknowledgement}
This paper is supported by the National Regional Innovationand Development Joint Fund (No. U24A20254).  Junbo Zhao is also supported by the NSFC under Grants (No. 62402424) and the Fundamental Research Funds for the Zhejiang Provincial Universities (226-2024-00049). Chen Zhao is supported by NYU Shanghai Center for Data Science.
\bibliography{custom}
\clearpage
\newpage
\appendix

\section{Details of Main Experiments}
\label{appendix: details of experiments}
\paragraph{Baselines.} 
We selected two basic retrieval methods as baselines for experimental comparison: the text-based approach BM25 \citep{bm25} and the widely used embedding-based approach Contriever \citep{contriever}. 
\textbf{BM25} is a probabilistic ranking model scoring documents based on term frequency, inverse document frequency, and document length normalization. 
\textbf{Contriever}, more precisely, the off-the-shelf Contriever is a dense retrieval method that leverages unsupervised contrastive learning to build vector representations for queries and documents. Contriever excels in zero-shot and domain-agnostic scenarios due to its robust pretraining approach.

For better comparison, we also introduce several state-of-the-art embedding-based retriever models, i.e. 
BGE \citep{bge_m3}\footnote{https://huggingface.co/BAAI/bge-m3}, 
gemma \citep{gemma}\footnote{https://huggingface.co/google/gemma-2-2b-it}, 
NV-Embed \citep{lee2024nv}\footnote{https://huggingface.co/nvidia/NV-Embed-v2}. 
These models are widely recognized for their superior performance. The primary performance advantages of these models stem from their training on more extensive and diverse datasets.

\section{More Details about Random Selection}
\label{appendix:random}
\begin{table*}[htbp]
\centering
\renewcommand{\arraystretch}{1.2} 
\setlength{\tabcolsep}{10pt}      
\resizebox{0.8\textwidth}{!}{
\begin{tabular}{llcccccc}
\toprule
\multicolumn{1}{c}{\multirow{2}{*}{\textbf{Generator}}} & \multicolumn{1}{c}{\multirow{2}{*}{\textbf{Retriever}}} & \multicolumn{3}{c}{\textbf{\popqa}}   & \multicolumn{3}{c}{\textbf{\entityqa}}                 \\ 
\cmidrule(lr){3-5} \cmidrule(lr){6-8}
\multicolumn{2}{c}{}      & \textbf{Long-tail} & \textbf{Infrequent} & \textbf{Frequent} & \textbf{Long-tail} & \textbf{Infrequent} & \textbf{Frequent} \\ \hline
\multirow{4}{*}{LLaMA-3 8B}         & Contriever         & 23.7                &   34.9             &      39.1         &     21.8          &      25.4           &        30.2         \\
                                    & FULL-random       & 23.3                &   34.9            &      42.1          &    22.6                &      27.8         &     33.7          \\
                                    & \ours-random               & 25.2                &   32.5            &      42.0       &    27.4               &      27.1          &     32.4          \\
                                    & \ours               & 29.8                &   34.7             &      41.8          &    30.1                &      31.6           &     33.9          \\ \hline
\bottomrule
\end{tabular}}
\captionsetup{justification=justified,singlelinecheck=false}
\caption{End-to-end QA exact match accuracy comparison for RAG systems with different retriever and LLaMA-3 8B on the \popqa and the \entityqa datasets. ``FULL-random'' means the retriever model trained on the examples randomly sampled from the same distribution as the data source. ``\ours-random'' means the retriever model trained on the examples randomly sampled from the same long-tail distribution as RPDR.}
\label{tab:random_compare}
\end{table*}

As mentioned earlier, RPDR-random is a model trained with randomly sampled data of the same size as RPDR, drawn from the same long-tail portion and using the same training hyperparameters. In this section, we additionally report its performance in the end-to-end setting.
Furthermore, we introduce another baseline, denoted as Full-random, where the same number of training examples are randomly sampled from the full distribution (i.e., not restricted to the long-tail) ~\citep{popqa,EntityQuestions} and used to train a model under the same setup.

The results in \Cref{tab:random_compare} show that sampling from the full distribution does not effectively enhance RAG performance in the long-tail setting; its performance is instead more similar to the original Contriever model. 
Moreover, performing random selection directly on long-tail data does not lead to performance improvements, which is consistent with the retriever's standalone results.
In contrast, our proposed RPDR data selection strategy better preserves the model's performance under the long-tail distribution, aligning with our prior hypothesis.

\section{Data Augmentation Scale and Model Performance}
\begin{figure}[h!]
    \centering
    \includegraphics[width=0.48\textwidth]{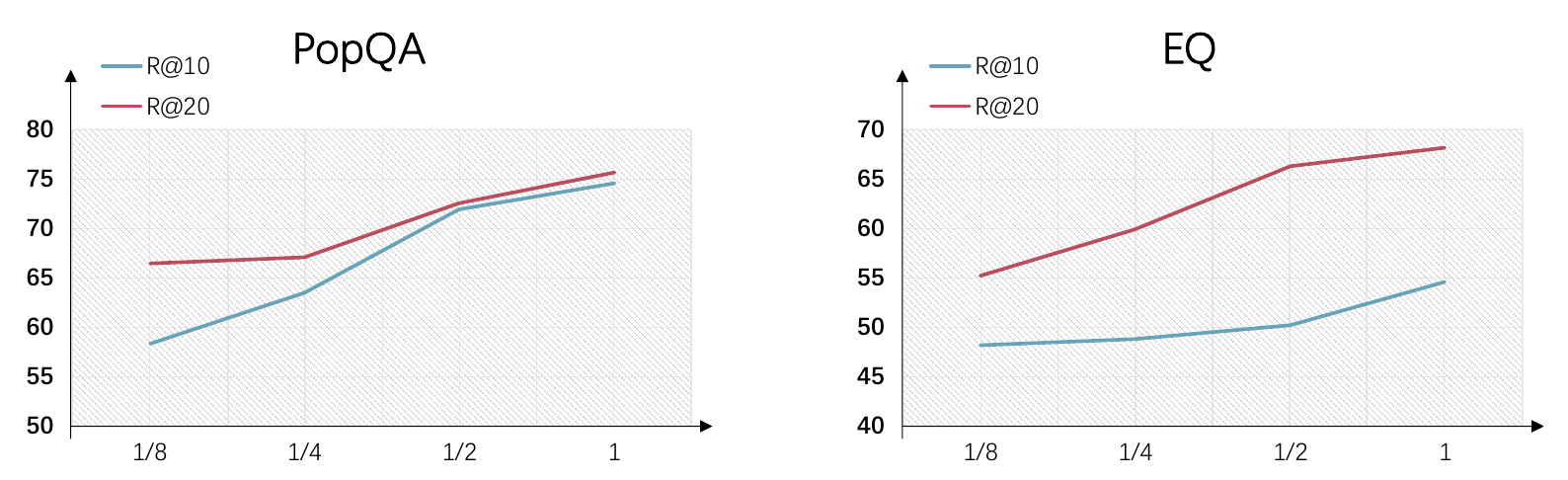}
    \captionsetup{justification=justified, singlelinecheck=false}  
    \caption{The Relationship between augmented data scale and retriever model performance. In the left figure, EQ represents \entityqa. The x-axis indicates the proportion of the dataset size relative to the full augmentated dataset.}
    \label{fig:scale and performance}
\end{figure}

We conduct an experiment on the relationship between data augmentation size and model performance. According to \Cref{fig:scale and performance}, the performance gradually increases with more augmented data, as expected.

\section{Main Results evaluated using IR Metric}
In the main body of this paper, we adopt the evaluation metric commonly used in prior work~\citep{sciavolino2021simple,wang2025proprag,yu2024rankrag,popqa}, \textbf{Exact Match}, which checks whether the gold answer appears in the retrieved passage. In this section, we additionally report standard metrics from the information retrieval (IR) domain. Specifically, we treat the first paragraph of the corresponding Wikipedia entity as the gold passage, following ~\citet{shavarani2024entity}. 

\begin{table*}[htbp]
\centering
\setlength{\tabcolsep}{10pt}      
\resizebox{0.9\textwidth}{!}{
\begin{tabular}{lccccccc}
\toprule
\multicolumn{1}{c}{\multirow{3}{*}{\textbf{Method}}}  & \multicolumn{3}{c}{\textbf{\popqa}}         & \multicolumn{3}{c}{\textbf{\entityqa}}         \\
\cmidrule(lr){2-4} \cmidrule(lr){5-7} 
\multicolumn{1}{c}{}  & \textbf{Long-tail} & \textbf{Infrequent} & \textbf{Frequent} & \textbf{Long-tail} & \textbf{Infrequent} & \textbf{Frequent} \\
\multicolumn{1}{c}{}  & Recall@10  & Recall@10  & Recall@10 & Recall@10 & Recall@10 & Recall@10 \\
\hline
BM 25     & 54.3  & 48.5  & 66.9  & 41.7  & 51.3 & 60.8   \\
Contriever       & 61.2 & 67.4  & 87.3 & 56.9 & 63.2 & 73.1 \\
BGE             & 66.4 & 77.5 & 89.7 & 61.1 & 69.8 & 77.1  \\
NV-Embed              & 71.8  & \textbf{80.8} & \textbf{94.5}  & 67.3 & 75.2 & \textbf{81.9} \\
Gemma             & 67.6  & 75.3 & 92.5 & 67.4 & 73.2  & 79.0 \\

\hline
\ours               & \textbf{74.3} & 79.1  & 88.3  & \textbf{72.6} & \textbf{75.4}  & 78.7 \\
\ours-Random                 & 69.2 & 73.6 & 88.9 & 68.5 & 74.3  & 77.4 \\
\bottomrule
\end{tabular}
}
\caption{Comparison of retrieval performance between RPDR and baseline retrievers for questions with long-
tail, infrequent, and frequent entities in the POPQA and ENTITYQUESTIONS datasets. ``Recall@10" means the gold passage being in the top-10 recall results.}
\label{tab:IR_metric}
\end{table*}

The detailed results are shown in Table~\ref{tab:IR_metric}. It can be clearly observed that the conclusion, ``RPDR significantly enhances long-tail retrieval", drawn from our main results, still holds under the modified metric. 
We also observe that dense retrievers tend to perform better in retrieving the gold passage, which may be attributed to the strong alignment between the training objective and the retrieval task.

However, we emphasize that the gold passages used in this evaluation were automatically constructed by extracting the first paragraph of the corresponding Wikipedia entity using scripts, without human annotation. Therefore, the results in this table should be considered for reference only.

\section{Full Dataset vs. Data After Filter}

\begin{table}[]
\resizebox{0.5\textwidth}{!}{
\begin{tabular}{lccc}
\hline
Model & Long-tail & Infrequent & Frequent \\
\hline
Contriever & 52.3 & 53.3 & 80.0 \\
RPDR-random & 66.8 & 70.0 & 82.7 \\
RPDR (Q wrong \& A wrong) & 62.9 & 53.9 & 65.8 \\
RPDR-trained with full dataset & 72.1 & 71.9 & 73.4 \\
RPDR & 74.5 & 76.8 & 81.3 \\
\hline
\end{tabular}}
\caption{Comparison across different training data.}
\label{tab:full_vs_filter}
\end{table}

Considering that the full dataset contains a larger volume of data, in this section we will discuss the necessity of the data filtering step. We report R@10 retrieval performance on PopQA in ~\Cref{tab:full_vs_filter}. The results show that training on the full set of 86k samples leads to suboptimal retrieval performance across long-tail, infrequent, and frequent queries. We suspect that the presence of noisy and low-quality synthetic data in the full dataset contributes to representation drift. Moreover, filtering reduces the training data size from 86k to 22k, significantly lowering computational costs. By comparing RPDR with baselines trained on randomly sampled data, wrongly generated data, and the full dataset, we confirm that RPDR offers advantages in both higher accuracy and lower cost.

\section{Scaling Dynamics of the Training Dataset over Multi-cycle Round Trips}
Would a second round-trip filtering cycle help the already fine-tuned model? We ran the procedure again and obtained fewer than 1k additional examples—too small compared with the original 22k pool—so we chose not to retrain the fine-tuned model with this marginal increment. ~\Cref{tab:scale_dynamic} shows the variation in training data size across multiple cycles.
\begin{table}[ht]
\begin{tabular}{lccc}
\hline
 & Cycle 1 & Cycle 2 & Cycle 3 \\
\hline
Data scale & 22k & 23k & 23k \\
\hline
\end{tabular}
\caption{Comparison across different training data.}
\label{tab:scale_dynamic}
\end{table}
Given the minor variation in scale, we refrained from extending the experiments to the second and third rounds.

\section{The Samples for Qualitative Snalysis}
As mentioned in \Cref{sec:qualitative}, we analyzed some correct and error case for qualitatively assessing the strengths and weaknesses of RPDR. In this part, we will present the specific samples that could not be included in the main text due to space limitations in \Cref{tab:error_cases_appendix}.
\label{appendix: Case Study}

\noindent\textbf{Hard-to-embed} means syntactically complex entities. These entities (\eg \underline{Ernő Noskó} often contain complex morphological structures or rare characters, making them harder to learn. Moreover, they are not included as augmented data (as our inverse embeddings also fail on them).

\noindent\textbf{Semantic Ambiguity} means entities with multiple meanings tend to be interpreted by the model based on their more frequently used semantics, causing it to over-prioritize common but incorrect matches.
For instance, terms like \textit{Milk} and \textit{Finale}, which have multiple plausible interpretations, are often associated with the retriever ranking popular but unrelated results above the correct but less common ones.

\noindent\textbf{Unanswerable Queries} means the query has no answer in the retrieved passages even though the description is truly related to the target entity. This type of error is primarily caused by issues in the design of sample answers. 

\begin{table*}[ht]
\centering
\renewcommand{\arraystretch}{1.5} 
\setlength{\tabcolsep}{8pt}      
\resizebox{\textwidth}{!}{%
\begin{tabular}{|p{2.5cm}|p{3.5cm}|p{2cm}|p{7cm}|}
\hline
\rowcolor[HTML]{EFEFEF} 
\textbf{Error Type} & \textbf{Query} & \textbf{Answer} & \textbf{Retrieved Passage} \\ \hline
Hard-to-Embed Target Entities & What is \textit{Edwin Wallock}'s occupation? & actor, \quad\quad actress & 
Charles Edwin (died 1756): The Government could field no candidates at the rerun of the election and Edwin was returned unopposed as Member of Parliament (MP) for Westminster ... Charles Edwin (died 1756) 
\\ \hline
Semantic \quad Ambiguity & What genre is \textit{Milk}? & biographical film, biopic & 
Milk: ... Freezing of milk can cause fat globule aggregation upon thawing, resulting in milky layers and butterfat lumps. ... Milk is often served in coffee and tea. Steamed milk is used to prepare espresso-based drinks such as cafe latte.
\\ \hline
Semantic \quad Ambiguity & Who was the director of \textit{Finale}? & Ken Kwapis, Kenneth William Kwapis & 
Finale (software): Finale is the flagship program of a series of proprietary music notation software developed and released by MakeMusic for the Microsoft Windows and macOS operating systems. ... including the score for an entire ensemble (e.g., orchestra, concert band, big band, etc.) and parts for the individual musicians.  
\\ \hline
Unanswerable Queries & What genre is \textit{Great Expectations}? & fiction, \quad\quad fictional & 
Great Expectations: The film adaptation in 1946 gained the greatest acclaim... Following are highlights of the adaptations for film and television, and for the stage, since the early 20th century. 
\\ \hline
\end{tabular}%
}
\captionsetup{justification=justified,singlelinecheck=false}
\caption{Three major categories of RPDR errors.}
\label{tab:error_cases_appendix}
\end{table*}

\end{document}